\documentclass[runningheads]{llncs}
\usepackage{graphicx}
\usepackage{cite}
\usepackage{graphicx, array, blindtext}
\begin{document}
\title{Iconographic Image Captioning for Artworks }
%
%
\author{Eva Cetinic\inst{1}\orcidID{0000-0002-5330-1259} }
\authorrunning{E. Cetinic}
%
\institute{Rudjer Boskovic Institute, Bijenicka cesta 54, 10000 Zagreb, Croatia
\email{ecetinic@irb.hr}\\
}
\maketitle              
\begin{abstract}
Image captioning implies automatically generating textual descriptions of images based only on the visual input. Although this has been an extensively addressed research topic in recent years, not many contributions have been made in the domain of art historical data. In this particular context, the task of image captioning is confronted with various challenges such as the lack of large-scale datasets of image-text pairs, the complexity of meaning associated with describing artworks and the need for expert-level annotations. This work aims to address some of those challenges by utilizing a novel large-scale dataset of artwork images annotated with concepts from the Iconclass classification system designed for art and iconography. The annotations are processed into clean textual description to create a dataset suitable for training a deep neural network model on the image captioning task. Motivated by the state-of-the-art results achieved in generating captions for natural images, a transformer-based vision-language pre-trained model is fine-tuned using the artwork image dataset. Quantitative evaluation of the results is performed using standard image captioning metrics. The quality of the generated captions and the model’s capacity to generalize to new data is explored by employing the model on a new collection of paintings and performing an analysis of the relation between commonly generated captions and the artistic genre. The overall results suggest that the model can generate meaningful captions that exhibit a stronger relevance to the art historical context, particularly in comparison to captions obtained from models trained only on natural image datasets.

\end{abstract}
\section{Introduction}

Automatically generating meaningful and accurate image descriptions is a challenging task that has been extensively addressed in the recent years. This task implies recognizing objects and their relationship in an image and generating syntactically and semantically correct textual descriptions. In resolving this task, significant progress has been made using deep learning based - techniques. A prerequisite for this kind of approach are large datasets of semantically related image and sentence pairs. In the domain of natural images, several well-known large-scale datasets are commonly used for caption generation, such as the MS COCO \cite{lin2014microsoft}, Flickr30 \cite{young2014image} and Visual Genome \cite{krishna2017visual} dataset. 
Although the availability of such datasets enabled remarkable results in generating high quality captions for photographs of various objects and scenes, the task of generating image captions still remains difficult for domain-specific image collections. In particular, in the 
context of the cultural heritage domain, generating image captions is an open problem with various challenges. One of the major obstacles is the lack of a truly large-scale dataset of artwork images paired with adequate descriptions. It is also relevant to address what kind of description would be regarded as "adequate" for a particular purpose. Considering for instance Erwin Panofsky’s three levels of analysis \cite{panofsky1972studies}, we can distinguish the "pre-iconographic" description, "iconographic" description and the "iconologic" interpretation as possibilities of aligning semantically meaningful, yet very different textual descriptions with the same image. While captions of natural images usually function on the level of "pre-iconographic" descriptions, which implies simply listing the elements that are depicted in an image, for artwork images this type of description represent only the most basic level of visual understanding and is often not considered to be of great interest. 
\newline
\indent In the context of artwork images, it would be more interesting to generate "iconographic" captions that capture the subject and symbolic relations between objects. Creating a dataset for such a complex task requires expert knowledge in the process of collecting sentence-based descriptions of images. There have been some attempts to create such datasets, but those existing datasets consist only of a few thousand images and are therefore not suitable to train deep neural models in the current state-of-the-art setting for image captioning. However, there are several existing large-scale artwork collections that associate images with keywords and specific concepts. The idea of this work is to use a concatenation of concept descriptions associated with an image as textual inputs for training an image captioning model. Recently an interesting large-scale artwork dataset has been published under the name "Iconclass AI Test Set" \cite{posthumus2020brill}. This dataset represents a collection of various artwork images assigned with alphanumeric classification codes that correspond to notations from the Iconclass system \cite{couprie1983iconclass}. Iconclass is a classification system designed for art and iconography and is widely accepted by museums and art institutions as a tool for the description and retrieval of subjects represented in images. Although the "Iconclass AI Test Set" is not structured primarily as an image captioning dataset, each code is paired with its "textual correlate" - a description of the iconographic subject of the particular Iconclass notation. Therefore the main intention of this work is to extract and preprocess the given annotations into clean textual description and create the "Iconclass Caption" dataset. This dataset is then used to fine-tune a pre-trained unified vision-language model on the down-stream task of image captioning \cite{zhou2020unified}. Transformer-based vision-language pre-trained models currently represent the leading approach in solving a variety of tasks in the intersection of computer vision and natural language processing. This paper represents a first attempt to employ the aforementioned approach on a collection of artwork images with the goal to generate image captions relevant in the context of art history.

\section{Related work}

The availability of large collections of digitized artwork images led to an increase of interest in the employment of deep learning-based techniques for a variety of different tasks. Research in this area most commonly focuses on addressing problems related to computer vision in the context of art historical data, such as image classification \cite{cetinic2018fine, sandoval2019two}, visual link retrieval \cite{seguin2016visual, castellano2020towards}, analysis of visual patterns and conceptual features \cite{shen2019discovering, deng2020exploring, cetinic2020learning, elgammal2018shape}, object and face detection \cite{crowley2014search, strezoski2018omniart}, pose and character matching \cite{madhu2019recognizing, jenicek2019linking} and computational aesthetics \cite {hayn2017subjective, cetinic2019deep, sargentis2020aesthetical}. 

Recently however there has been a surge of interest in topics that deal with not only visual, but both visual and textual modalities of artwork collections. The pioneering works in this research area mostly addressed the task of multi-modal retrieval. In particular, \cite{garcia2018read} introduced the SemArt dataset, a collection of fine-art images associated with textual comments, with the aim to map the images and their descriptions in a joint semantic space. They compare different combinations of visual and textual encodings, as well as different methods of multi-modal transformation. In projecting the visual and textual encodings in a common multimodal space, they achieve the best results by applying a neural network trained with cosine margine loss on ResNet50 features as visual encodings and bag-of-word as textual encodings. The task of creating a shared embedding space was also addressed in \cite{baraldi2018aligning} where the authors introduce a new visual semantic dataset named BibleVSA, a collection of miniature illustrations and commentary text pairs, and explore supervised and semi-supervised approaches to learning cross-references between textual and visual information in documents. In \cite{stefanini2019artpedia} the authors present the Artpedia dataset consisting of 2930 images annotated with visual and contextual sentences. They introduce a cross-modal retrieval model that projects images and sentences in a common embedding space and discriminates between contextual and visual sentences of the same image. A similar extension of this approach to other artistic datasets was presented in \cite{cornia2020explaining}.
 
Besides multi-modal retrieval, another emerging topic of interest is visual question answering (VAQ). In \cite{bongini2020visual} the authors annotated a subset of the ArtPedia dataset with visual and contextual question-answer pairs and introduced a question classifier that discriminates between visual and contextual questions and a model that is able to answer both types of questions. In \cite{garcia2020dataset} the authors introduce a novel dataset AQUA, which consists of automatically generated visual and knowledge-based QA pairs, and also present a two-branch model where the visual and knowledge questions are handled independently.  

A limited number of studies contributed to the task of generating descriptions of artwork images using deep neural networks and all of them rely on employing the encoder-decoder architecture-based image captioning approach. For example, \cite{sheng2019generating} proposes an encoder-decoder framework for generating captions of artwork images where the encoder (ResNet18 model) extracts the input image feature representation and the artwork type representation, while the decoder is a long short-term memory (LSTM) network. They introduce two image captioning datasets referring to ancient Egyptian art and ancient Chinese art, which contain 17,940 and 7,607 images respectively. Another very recent work \cite{guptatowards} presented a novel captioning dataset for art historical images consisting of 4000 images across 9 iconographies, along with a description for each image consisting of one or more paragraphs. They used this dataset to fine-tune different variations of image captioning models based on the well-known encoder-decoder approach introduced in \cite{vinyals2015show}. 
 
Influenced by the success of utilizing large scale pre-trained language models like BERT \cite{devlin2018bert} for different tasks related to natural language processing, there has recently been a surge of interest in developing Transformer-based vision-language pre-trained models. Vision-language models are designed to learn joint representations that combine information of both modalities and the alignments across those modalities. It has been shown that models pre-trained on intermediate tasks with unsupervised learning objectives using large datasets of image-text pairs, achieve remarkable results when adapted to different down-stream tasks such as image captioning, cross-modal retrieval or visual question answering \cite{zhou2020unified, tan2019lxmert, lu2019vilbert, chen2019uniter}. However, to the best of our knowledge, this approach has until now not been explored for tasks in the domain of art historical data.

\section{Experimental setup}

\subsection{Iconclass Caption Dataset} 

\begin{figure}[!h]
\begin{minipage}{0.2\textwidth}
	\includegraphics[width=2.5cm,height=3.5cm]{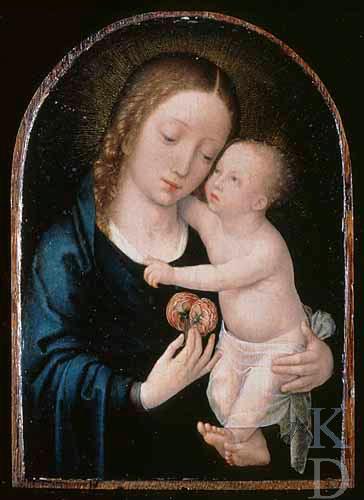}
\end{minipage}
\hspace{0.2cm}
\begin{minipage}{0.8\textwidth}\raggedright
	{\fontsize{8}{10} \textbf {Original description:} Madonna: i.e. Mary with the Christ-child, flowers: rose, historical persons (portraits and scenes from the life) (+ half-length portrait) 
	\newline
	\textbf {Clean description:} Madonna: i.e. Mary with the Christ-child, flowers: rose, historical persons .}
\end{minipage}

\vspace{0.2cm}
\begin{minipage}{0.2\textwidth}
	\includegraphics[width=2.5cm,height=3.5cm]{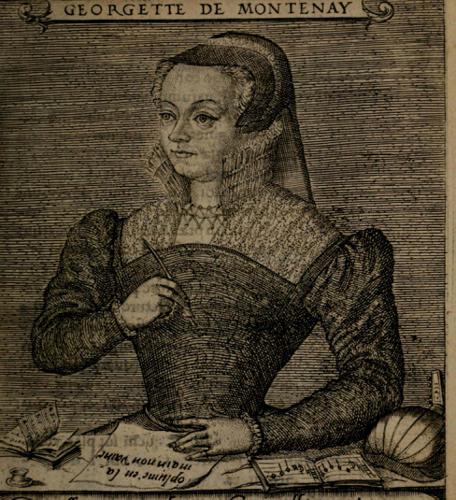}
\end{minipage}
\hspace{0.2cm}
\begin{minipage}{0.8\textwidth}\raggedright
	{\fontsize{8}{10} \selectfont \textbf {Original description:} adult woman, manuscript of musical score, writer, poet, author (+ portrait, self-portrait of artist), pen, ink-well, paper (writing material), codex, inscription, historical events and situations (1567), historical person (MONTENAY, Georgette de) - BB - woman - historical person (MONTENAY, Georgette de) portrayed alone, proverbs, sayings, etc. (O PLUME EN LA MAIN NON VAINE) 
	\newline
	\textbf {Clean description:} adult woman, manuscript of musical score, writer, poet, author , pen, ink-well, paper , codex, inscription, historical events and situations , historical person, woman - historical person portrayed alone, proverbs, sayings.}
\end{minipage}

\vspace{0.2cm}
\begin{minipage}{0.2\textwidth}
	\includegraphics[width=2.5cm,height=3.5cm]{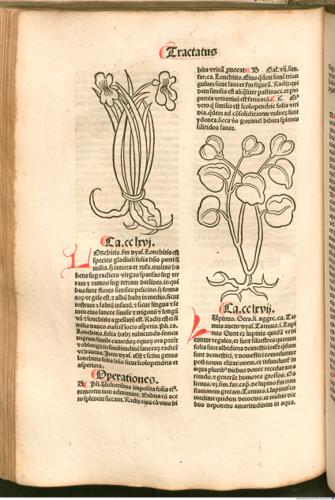}
\end{minipage}
\hspace{0.2cm}
\begin{minipage}{0.8\textwidth}\raggedright
	{\fontsize{8}{10}\textbf {Original description:} plants and herbs (HELLEBORINE), plants and herbs (LUPINE), 
	\newline
	\textbf {Clean description:} plants and herbs .}
\end{minipage}

\caption{Example images from the Iconclass dataset and their corresponding descriptions before and after preprocessing.} \label{fig1}

\end{figure}

In our experiment we use a subset of 86 530 valid images from the "Iconclass AI Test Set" \cite{posthumus2020brill}.This is a very diverse collection of images sampled from the Arkyves database \footnote{www.arkyves.org}. It includes images of various types of artworks such as paintings, posters, drawings, prints, manuscripts pages, etc. Each image is associated with one or more codes linked to labels from the Iconclass classification system. The authors of the "Iconclass AI Test Set" provide a json file with the list of images and corresponding codes, as well as an Iconclass Python package to perform analysis and extract information from the assigned classification codes. To extract textual descriptions of images for the purpose of this work, the English textual descriptions of each code associated with an image are concatenated. Further preprocessing of the descriptions includes removing text in brackets and some recurrent uppercased dataset-specific codes. In this dataset, the text in brackets most commonly includes very specific named entities, which are considered a noisy input in the image captioning task. Therefore, when preprocessing the textual items, all the text in brackets is removed, even at the cost of sometimes removing useful information. Figure \ref{fig1} shows several example images from the Iconclass dataset and their corresponding descriptions before and after preprocessing. Depending on the number of codes associated with each image, the final textual descriptions can significantly vary in length. Also, because of the specific properties of this dataset, the image descriptions are not structured as sentences but as a list of comma-separated words and phrases. 

Because of this type of structure, and because of having only one reference caption for each image, the Iconclass Caption dataset is not a standard image captioning dataset. However, having in mind the difficulties of obtaining adequate textual descriptions for images of artworks, this dataset can be considered a valuable source of image-text pairs in the current context. Particularly because of the large number of annotated images that enables training deep neural models. In the experimental setting, a subset of 76k items is used for training the model, 5k for validation and 5k for testing.

\subsection{Vision-Language Model}

In this work the unified vision-language pre-training model (VLP) introduced in \cite{zhou2020unified} is employed. This model is denoted as “unified” because the same pre-trained model can be fine-tuned for different types of tasks. Those task include both vision-language generation (e.g. image captioning) and vision-language understanding (e.g. visual question answering). The model is based on an encoder-decoder architecture comprised of 12 Transformer blocks. The model input consist of the image embedding, text embedding and three special tokens that indicate the start of the image input, the boundary between visual and textual input and the end of the textual input. The image input consist of 100 object classification aware region features extracted using the Faster RCNN model \cite{ren2015faster} pre-trained on the Visual Genome dataset \cite{krishna2017visual}. For a more detailed description of the overall VLP framework and pre-training objectives, the reader is refered to \cite{zhou2020unified}. The experiments introduced in this work employ as the base model the VLP model pre-trained on the Conceptual Captions dataset \cite{sharma2018conceptual} using the sequence-to-sequence objective. This base model is fine-tuned on the Iconclass Caption Dataset using recommended fine-tuning configurations, namely training with a constant learning rate of 3e-5 for 30 epochs. Because the descriptions in the  Iconclass Caption Dataset are on average longer than captions in other caption datasets, when fine-tuning the VLP model,  the maximum number of tokens in the input and target sequence is modified from the default value (20) to a new higher value (100).

\section{Results}

\subsection{Quantitative results}

To quantitatively evaluate the generated captions, standard language evaluation metrics for image captioning on the Iconclass Caption test set are used. Those include the standard 4 BLEU metrics \cite{papineni2002bleu}, METEOR \cite{denkowski2014meteor} ROUGE \cite{lin2004rouge} and CIDEr \cite{vedantam2015cider}. The BLUE, ROUGE and METEOR are metrics that originate from machine translation tasks, while CIDEr was specifically developed for image caption evaluation. The BLUE metrics represent n-gram precision scores multiplied by a brevity penalty factor to assess the length correspondence of candidate and reference sentences. ROUGE is a metric that measures the recall of n-grams and therefore rewards long sentences. Specifically ROUGE-L measures the longest matching sequence of words between a pair of sentences. METEOR represents the harmonic mean of precision and recall of unigram matches between sentences and additionally  includes synonyms and paraphrase matching. CIDEr measures the cosine similarity between TF-IDF weighted n-grams of the candidate and the reference sentences. The TF-IDF weighting of n-grams reduces the score of frequent n-grams and appoints higher scores to distinctive words. The results obtained using those metrics are presented in Table~\ref{tab1}.

\begin{table}
\centering
\caption{Table captions should be placed above the
tables.}\label{tab1}
\begin{tabular}{|l|c|}
\hline
Evaluation metric &  Iconclass Caption test set\\
\hline
BLEU 1 & $14.8 $ \\
BLEU 2 & $12.8 $\\
BLEU 3 & $11.3 $\\
BLEU 4 & $10.0 $\\
METEOR & $11.7 $\\
ROUGE-L & $31.9 $\\
CIDEr & $172.1 $\\

\hline
\end{tabular}
\end{table}

Although the current results cannot be compared with any other work because the experiments are performed on a new and syntactically and semantically different dataset, the quantitative evaluation results are included to serve as a benchmark for future work. In comparison with current state-of-the-art caption evaluation results on natural image datasets (e.g. BLEU4 $\approx$ 37 for COCO  and $\approx$ 30 for Flickr30 datasets) \cite{xia2020xgpt, zhou2020unified}, the BLUE scores are lower for the Iconclass dataset. A similar behaviour is also reported in another study addressing iconographic image captioning \cite{guptatowards}. On the other hand, the CIDEr score is quite high in comparison to the one reported for natural image datasets (e.g. CIDEr $\approx$ 116 for COCO and  $\approx$ 68 for Flickr30 dataset) \cite{xia2020xgpt, zhou2020unified}. 

\begin{figure}[!h]
	\centering
	\begin{minipage}{0.2\textwidth}
		\includegraphics[width=2.2cm,height=2.7cm]{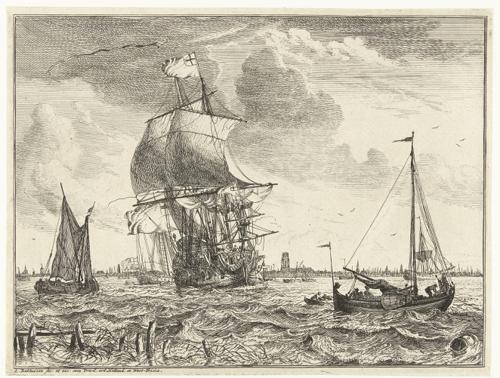}
	\end{minipage}
	\hspace{0.2cm}
	\begin{minipage}{0.45\textwidth}\raggedright
		{\fontsize{9}{10}\selectfont \textbf {Ground-truth:} sea.
			\newline
			\textbf {Caption:} sailing - ship, sailing - boat.}
	\end{minipage}
	\hspace{0.2cm}
	\begin{minipage}{0.2\textwidth}\raggedright
	{\fontsize{6}{10}\selectfont	 BLEU 1: 2.49e-16 \\
		BLEU 2: 2.88e-16 \\
		BLEU 3: 3.46e-16 \\
		BLEU 4: 4.51e-16 \\
		METEOR: 0.16 \\
		ROUGE: 0.0 \\
		CIDEr: 0.0}
	\end{minipage}
	\vspace{0.2cm}
	
	\begin{minipage}{0.2\textwidth}
	\includegraphics[width=2.2cm,height=2.7cm]{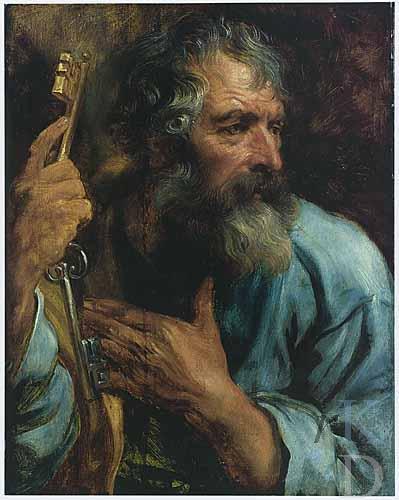}
	\end{minipage}
	\hspace{0.2cm}
	\begin{minipage}{0.45\textwidth}\raggedright
		{\fontsize{9}{10}\selectfont \textbf {Ground-truth:} apostle, unspecified, key.
			\newline
			\textbf {Caption:} head turned to the right, historical persons.}
	\end{minipage}
	\hspace{0.2cm}
	\begin{minipage}{0.2\textwidth}\raggedright
		{\fontsize{6}{10}\selectfont BLEU 1: 1.43e-16 \\
			BLEU 2: 1.54e-16 \\
			BLEU 3: 1.68e-16 \\
			BLEU 4: 1.85e-16 \\
			METEOR: 0.0 \\
			ROUGE: 0.0 \\
			CIDEr: 0.0}
	\end{minipage}
	\vspace{0.2cm}
	
	\begin{minipage}{0.2\textwidth}
		\includegraphics[width=2.2cm,height=2.7cm]{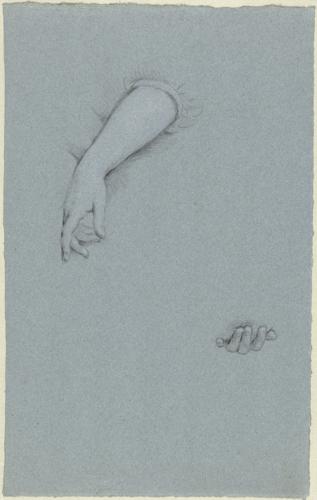}
	\end{minipage}
	\hspace{0.2cm}
	\begin{minipage}{0.45\textwidth}\raggedright
		{\fontsize{9}{10}\selectfont \textbf {Ground-truth:} arms, fingers.
			\newline
			\textbf {Caption:} hand.}
	\end{minipage}
	\hspace{0.2cm}
	\begin{minipage}{0.2\textwidth}\raggedright
		{\fontsize{6}{10}\selectfont BLEU 1: 3.67e-16 \\
			BLEU 2: 1.16e-11 \\
			BLEU 3: 3.67e-10 \\
			BLEU 4: 2.06e-09 \\
			METEOR: 0.0 \\
			ROUGE: 0.0 \\
			CIDEr: 0.0}
	\end{minipage}
	
	\vspace{0.2cm}
	
	\begin{minipage}{0.2\textwidth}
	\includegraphics[width=2.2cm,height=2.7cm]{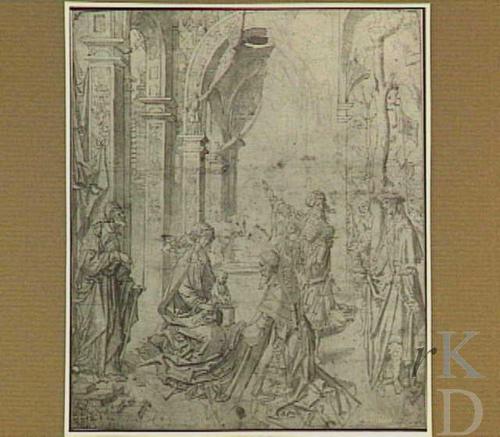}
	\end{minipage}
	\hspace{0.2cm}
	\begin{minipage}{0.45\textwidth}\raggedright
		{\fontsize{9}{10}\selectfont \textbf {Ground-truth:} palace, king, New Testament, adoration of the kings: the Wise Men present their gifts to the Christ-child.
			\newline
			\textbf {Caption:} New Testament.}
	\end{minipage}
	\hspace{0.2cm}
	\begin{minipage}{0.2\textwidth}\raggedright
		{\fontsize{6}{10}\selectfont BLEU 1: 0.00055 \\
			BLEU 2: 0.00055 \\
			BLEU 3: 5.53e-16 \\
			BLEU 4: 5.53e-07 \\
			METEOR: 0.0552 \\
			ROUGE: 0.184 \\
			CIDEr: 0.051}
	\end{minipage}
	
\caption{Examples of images from the Iconclass Caption test set, their corresponding ground-truth and generated captions and the values of evaluation metrics for those examples. } \label{fig2}
	
\end{figure}

However, it remains questionable how adequate these metrics are in assessing the overall quality of the captions in this particular context. All of the reported metrics mostly measure the word overlap between generated and reference captions. They are not designed to capture the semantic meaning of a sentence and therefore often lead to poor correlation with human judgement. Also, they are not appropriate for measuring very short descriptions which are quite common in the IconClass Caption dataset. Moreover, they do not address the relation between the generated caption and the image content, but express only the similarity between the original and generated textual descriptions. The generated caption could be semantically aligned with the image content but represent a different version of the original caption and therefore have very low metric scores. In Figure \ref{fig2}, several such examples from the Iconclass Caption test set are presented. 

Those examples indicate that  the existing evaluation metrics are not very suitable in assessing the relevance of generated captions for this particular dataset. Therefore a qualitative analysis of the results is also required in order to better understand potential contributions and drawbacks of the proposed approach. 

\subsection{Qualitative analysis}

For the purpose of qualitative analysis, examples of images and generated captions on two datasets are analyzed. One is the test set of the Iconclass Caption dataset that serves for direct comparison between the generated captions and ground-truth descriptions. The other dataset is a subset of the WikiArt painting collection, which does not include textual descriptions of images but has a broad set of labels associated with each image. This enables the study of the relation between generated captions and other concepts, e.g genre categorization of paintings, as well as gives an insight into how well the model generalizes to a different artwork dataset.

\subsubsection{Iconclass Caption test set}

\begin{figure}[!h]
	\centering
	\begin{minipage}{0.15\textwidth}
		\includegraphics[width=2cm,height=2.5cm]{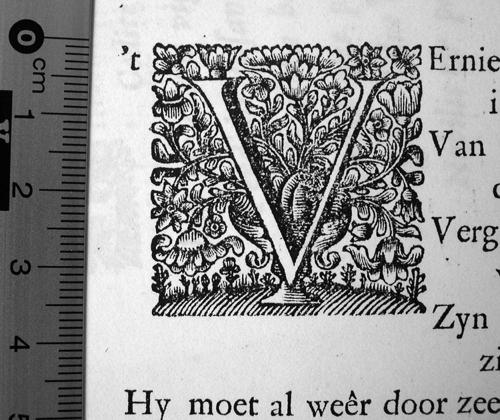}
	\end{minipage}
	\hspace{0.2cm}
	\begin{minipage}{0.3\textwidth}\raggedright
		{\fontsize{7}{7}\selectfont \textbf {Ground-truth:} historiated initial , printed historiated initial , printed matter , Roman script ; scripts based on the Roman alphabet
			\newline
			\textbf {Caption:} historiated initial , printed historiated initial , printed matter , Roman script ; scripts based on the Roman alphabet}
	\end{minipage}
	\hspace{0.2cm}
		\begin{minipage}{0.15\textwidth}
		\includegraphics[width=2cm,height=2.5cm]{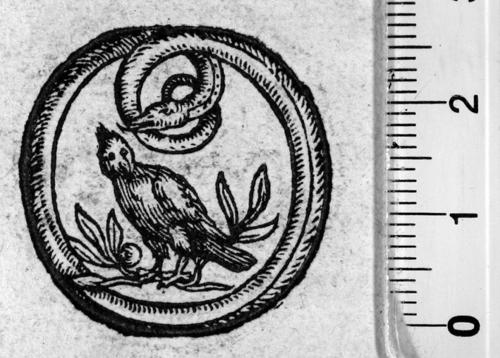}
	\end{minipage}
	\hspace{0.2cm}
	\begin{minipage}{0.3\textwidth}\raggedright
		{\fontsize{7}{7}\selectfont \textbf {Ground-truth:} device , printed matter.
			\newline
			\textbf {Caption:} historiated initial , printed historiated initial , printed matter , Roman script ; scripts based on the Roman alphabet.}
	\end{minipage}
	\vspace{0.2cm}
	
	\begin{minipage}{0.15\textwidth}
		\includegraphics[width=2cm,height=2.5cm]{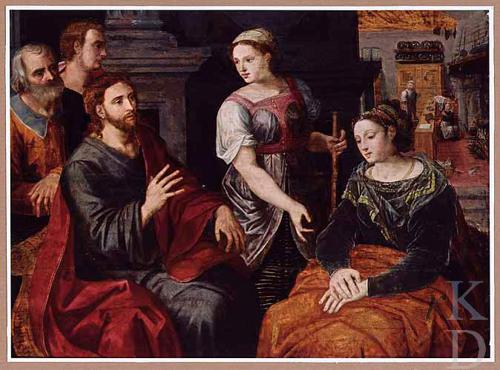}
	\end{minipage}
	\hspace{0.2cm}
	\begin{minipage}{0.3\textwidth}\raggedright
		{\fontsize{7}{7}\selectfont \textbf {Ground-truth:} 'New Testament, Christ in the house of Martha and Mary , Martha complains to Christ about Mary.
			\newline
			\textbf {Caption:} New Testament. }
	\end{minipage}
	\hspace{0.2cm}
	\begin{minipage}{0.15\textwidth}
		\includegraphics[width=2cm,height=2.5cm]{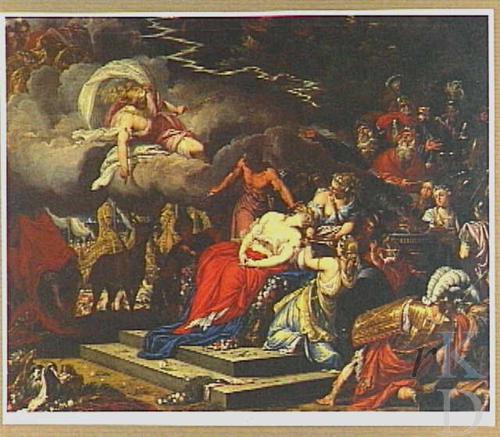}
	\end{minipage}
	\hspace{0.2cm}
	\begin{minipage}{0.3\textwidth}\raggedright
		{\fontsize{7}{7}\selectfont \textbf {Ground-truth:} hoofed animals: deer, Classical Mythology and Ancient History, the sacrifice of Iphigenia.
			\newline
			\textbf {Caption:} New Testament , the assumption of Mary : she is borne into heaven by angels.}
	\end{minipage}
	\vspace{0.2cm}
	
	\begin{minipage}{0.15\textwidth}
	\includegraphics[width=2cm,height=2.5cm]{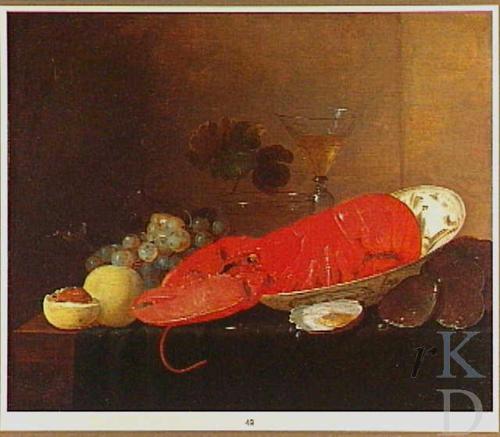}
	\end{minipage}
	\hspace{0.2cm}
	\begin{minipage}{0.3\textwidth}\raggedright
		{\fontsize{7}{7}\selectfont \textbf {Ground-truth:} 'crustaceans: lobster, molluscs: oyster, glass, rummer, table-cloth.
			\newline
			\textbf {Caption:} crustaceans : lobster , molluscs : oyster , fruits : lemon , fruits : orange , fruits : peach , fruits : walnut , table - cloth , fruit , fruit , grape . }
	\end{minipage}
	\hspace{0.2cm}
	\begin{minipage}{0.15\textwidth}
		\includegraphics[width=2cm,height=2.5cm]{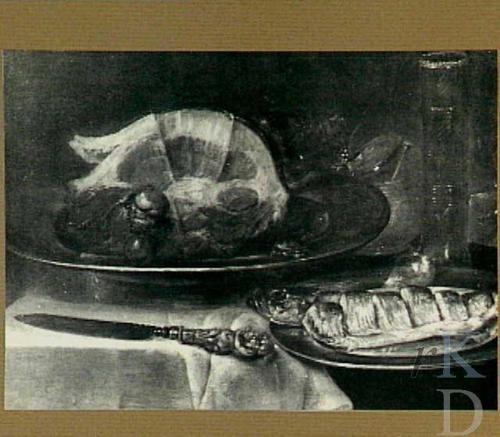}
	\end{minipage}
	\hspace{0.2cm}
	\begin{minipage}{0.3\textwidth}\raggedright
		{\fontsize{7}{7}\selectfont \textbf {Ground-truth:} glass, rummer, meat; still life of meat, tools, aids, implements ~ crafts and industries: knife.
			\newline
			\textbf {Caption:} crustaceans : lobster , molluscs : oyster , fish ; still life of fishes.}
	\end{minipage}
	\vspace{0.2cm}

	\begin{minipage}{0.15\textwidth}
		\includegraphics[width=2cm,height=2.5cm]{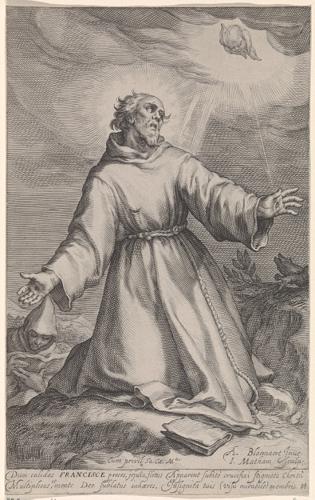}
	\end{minipage}
	\hspace{0.2cm}
	\begin{minipage}{0.3\textwidth}\raggedright
		{\fontsize{7}{7}\selectfont \textbf {Ground-truth:} 'founder of the Order of Friars Minor , Francisof Assisi; possible attributes: book, crucifix, lily, skull, stigmata.
			\newline
			\textbf {Caption:} male saints. }
	\end{minipage}
	\hspace{0.2cm}
	\begin{minipage}{0.15\textwidth}
		\includegraphics[width=2cm,height=2.5cm]{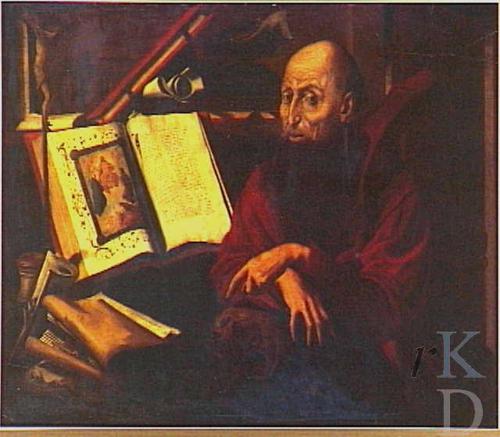}
	\end{minipage}
	\hspace{0.2cm}
	\begin{minipage}{0.3\textwidth}\raggedright
		{\fontsize{7}{7}\selectfont \textbf {Ground-truth:} saints, St. Jerome as Doctor of the Latin Church in his study with book, pen and ink; lion and cardinal's hat beside him, study; 'studiolo'; library.
			\newline
			\textbf {Caption:} saints , the penitent harlot Mary Magdalene ; possible attributes : book , crown , crown of thorns , crucifix , jar of ointment , mirror , musical instrument , palm - branch , rosary , scourge , book.}
	\end{minipage}
	\caption{Examples of images from the Iconclass Caption test set, their corresponding ground-truth and generated captions. Examples shown in the left column represent successfully generated captions, while examples shown in the right column demonstrate wrongly generated captions. } \label{fig3}

\end{figure}

To gain a better insight into the generated image captions, in Figure \ref{fig3} several examples are shown. The presented image-text pairs are chosen to demonstrate both successful examples (the left column) and failed examples (the right column) of generated captions.

Analysis of the failed examples indicates an existing “logic” in those erroneous captions, as well as demonstrates underlying biases within the dataset. For instance, in the Iconclass Caption training test there are more than thousand examples that include the phrase “New Testament” in the description. Therefore images that include structurally similar scenes, particularly from classical history and mythology, are sometimes wrongly attributed as depicting a scene from the New Testament. This signifies the importance of balanced examples in the training dataset and indicates directions for possible future improvements. The Iconclass dataset is a collection of very diverse images and apart from the Iconclass classification codes, there are currently no other metadata available for the images.  Therefore it is difficult to perform an in-depth exploratory analysis of the dataset and the generated results in regard to attributes relevant in the context of art history such as the date of creation, style, genre, etc. For this reason, the fine-tuned image captioning model is employed on a novel artwork dataset - a subset of the WikiArt collection of paintings.

\subsubsection{WikiArt dataset}

\begin{figure}[!h]
	\centering
	\includegraphics[width=0.99\textwidth, width=0.6\textheight]{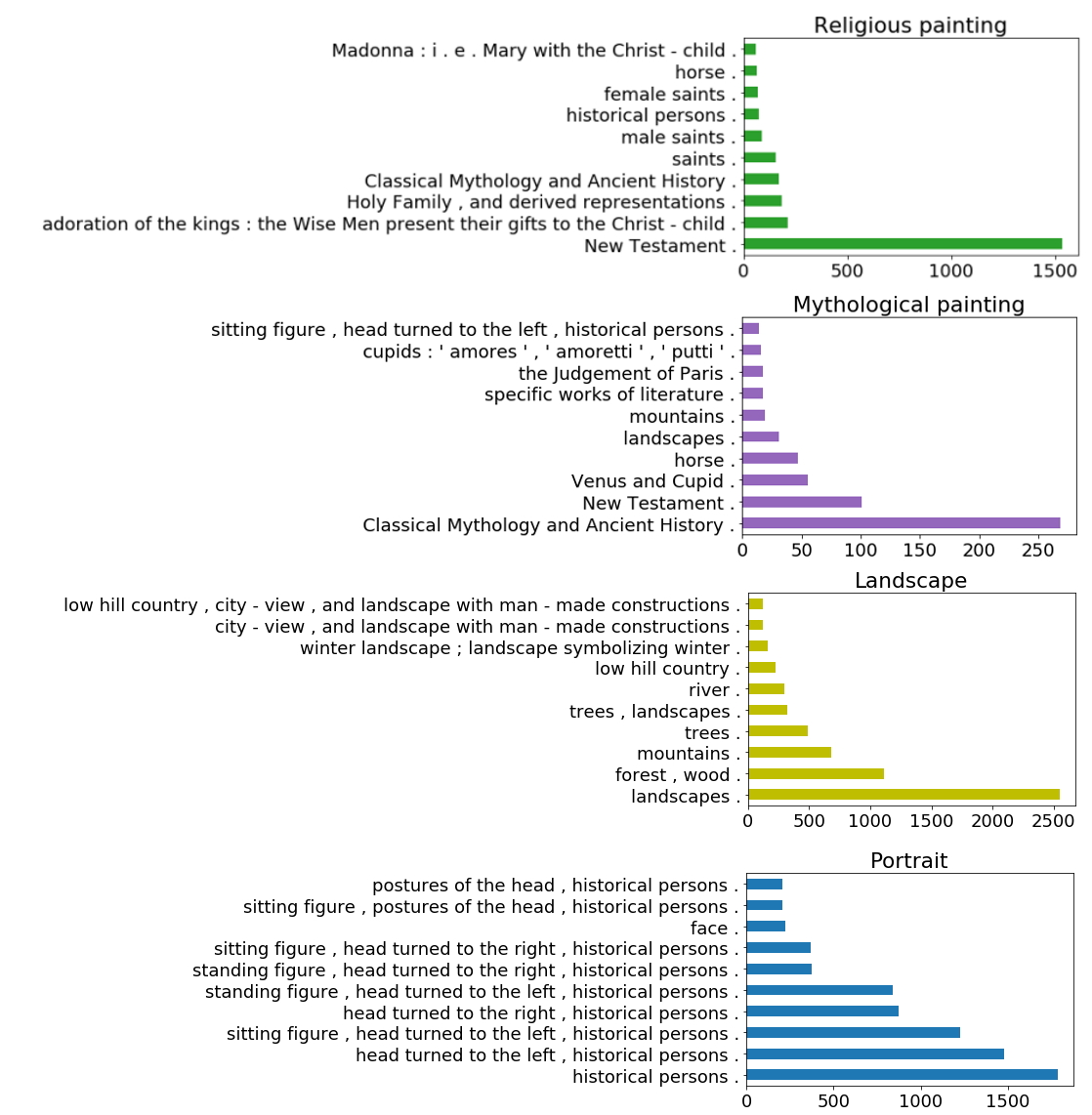}
	
	\caption{Distribution of most commonly generated descriptions in relation to four different genres in the WikiArt dataset.} \label{fig4}
\end{figure}

In order to explore how the model generalizes to a new artwork dataset, a subset of 52562 images of paintings from the WikiArt \footnote{www.wikiart.org} collection is used. Because images in the WikiArt dataset are annotated with a broad set of labels (e.g. style, genre, artist, technique, date of creation, etc. ), the study of the relation between the generated captions on those labels is performed as one method of qualitative assessment. Figure \ref{fig4} shows the distribution of most commonly generated descriptions in relation to four different genres. From this basic analysis it is obvious that the generated captions are meaningful in relation to the content and the genre classification of images.

\begin{figure}[!h]
	\centering
	\begin{minipage}{0.2\textwidth}
		\includegraphics[width=2cm,height=2.5cm]{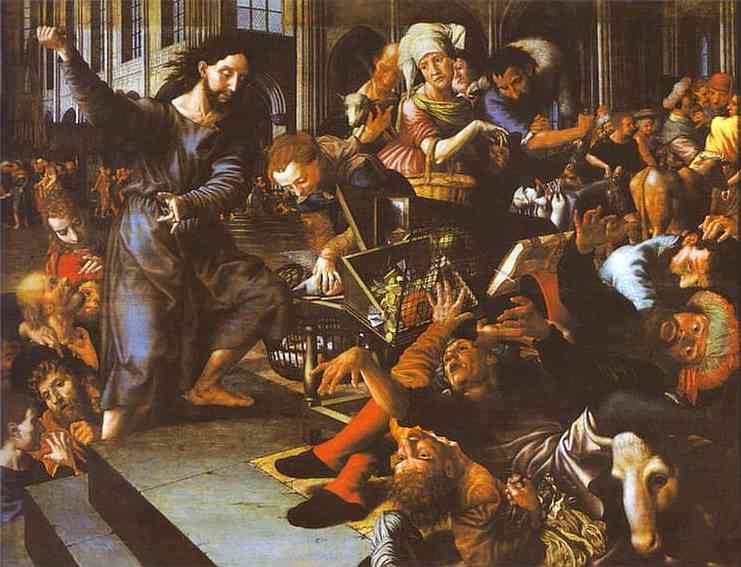}
	\end{minipage}
	\hspace{0.2cm}
	\begin{minipage}{0.75\textwidth}\raggedright
		{\fontsize{9}{10}\selectfont \textit {Jan van Hemessen, Christ Driving Merchants from the Temple, 1556}
			\newline
			\textbf {Iconclass caption:} New Testament .
			\newline
			\textbf {Flickr caption:}  A painting of a group of people .
		\newline
		\textbf {Coco caption:} A painting of a group of people dancing .}
	\end{minipage}

	\vspace{0.2cm}
	
		\centering
	\begin{minipage}{0.2\textwidth}
		\includegraphics[width=2cm,height=2.5cm]{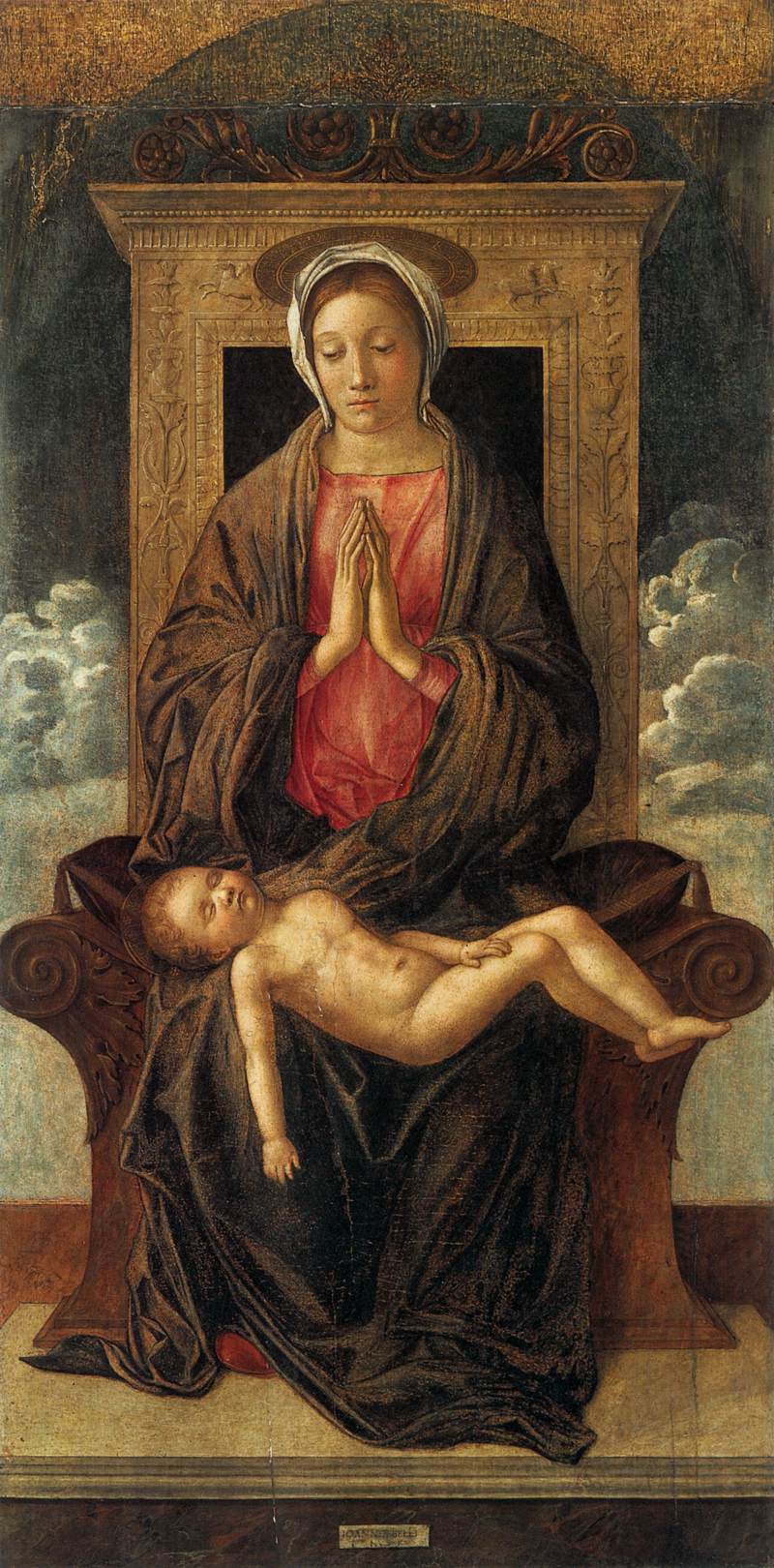}
	\end{minipage}
	\hspace{0.2cm}
	\begin{minipage}{0.75\textwidth}\raggedright
		{\fontsize{9}{10}\selectfont \textit {Giovanni Bellini, Madonna Enthroned Cherishing the Sleeping Child, 1475}
			\newline
			\textbf {Iconclass caption:} Madonna : i . e . Mary with the Christ - child , sitting figure , historical persons .
			\newline
			\textbf {Flickr caption:}  A woman holding a baby .
			\newline
			\textbf {Coco caption:} A painting of a woman holding a child .}
	\end{minipage}
   	\vspace{0.2cm}
   
   \centering
   \begin{minipage}{0.2\textwidth}
   	\includegraphics[width=2cm,height=2.5cm]{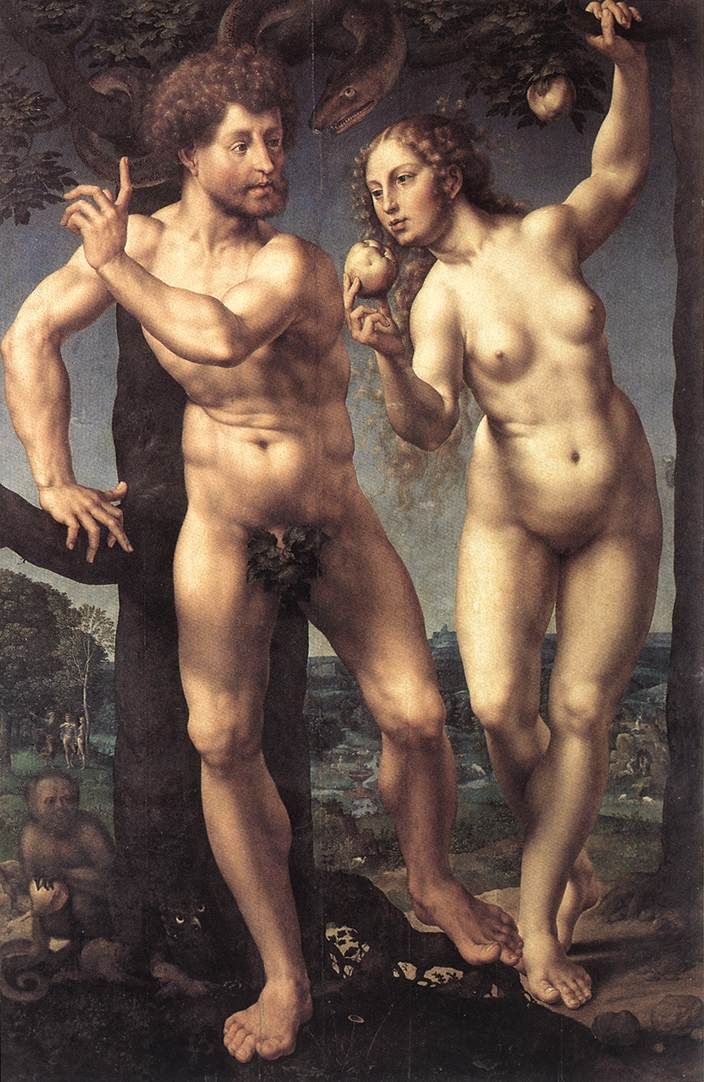}
   \end{minipage}
   \hspace{0.2cm}
   \begin{minipage}{0.75\textwidth}\raggedright
   	{\fontsize{9}{10}\selectfont \textit {Jan Gossaert, Adam and Eve in Paradise, 1527}
   		\newline
   		\textbf {Iconclass caption:} Adam and Eve holding the fruit .
   		\newline
   		\textbf {Flickr caption:}  Four naked men are standing in the mud .
   		\newline
   		\textbf {Coco caption:} A couple of men standing next to each other .}
   \end{minipage}

	\caption{Examples from the WikiArt dataset with captions generated by models fine-tuned on the Iconclass, Flickr and COCO datasets. } \label{fig5}
\end{figure}

To understand the contribution of the proposed model in the context of iconographic image captioning, it is interesting to compare the Iconclass captions with captions obtained from models trained on natural images. For this purpose, two models of the same architecture but fine-tuned on the Flickr 30 i MS COCO datasets are used. Figure \ref{fig5} shows several examples from the WikiArt dataset with corresponding Iconclass, Flickr and COCO captions. It is evident that the other two models generate results that are meaningful in relation to the image content but do not necessarily contribute to producing more fine-grained and context-aware descriptions. 

\section{Conclusion}

This paper introduces a novel model for generating iconographic image captions. This is done by utilizing a large-scale dataset of artwork images annotated with concepts from the Iconclass classification system designed for art and iconography. To the best of our knowledge, this dataset has not yet been widely used in the computer vision community. Within the scope of this work, the available annotations are processed into clean textual descriptions and the existing dataset is transformed into a collection of suitable image-text pairs. The dataset is used to fine-tune a transformer-based visual-language model. For this purpose, object classification aware region features are extracted from the images using the Faster RCNN model. The base model in our fine-tuning experiment is an existing model, called the VLP model, that is pre-trained on a natural image dataset on an intermediate tasks with unsupervised learning objectives. Fine-tuning pre-trained vision-language models represents the current state-of-the-art approach for many different multimodal tasks.   

The captions generated by the fine-tuned models are evaluated using standard image captioning metrics. Unlike in other image captioning datasets which usually contain several short sentences, the ground-truth descriptions of the Iconclass dataset significantly vary in length.  Because of the specific properties of the Iconclass dataset, standard image captioning evaluation metrics are not very informative regarding the relevance and appropriateness of the generated captions in relation to the image content. Therefore, the quality of the generated captions and the model’s capacity to generalize to new data are further explored by employing the model on another artwork dataset. The overall quantitative and qualitative evaluation of the results suggests that the model can generate meaningful captions that capture not only the depicted objects but also the art historical context and relation between subjects. However, there is still room for significant improvement. In particular, the unbalanced distribution of themes and topics within the training set result in often wrongly identified subjects in the generated image descriptions. Furthermore, the generated textual descriptions are often very short and could serve more as labels rather than captions. Nevertheless, the current results show significant improvement in comparison to captions generated from artwork images using models trained on natural image caption datasets. Further improvement can potentially be achieved with fine-tuning the current model on a smaller dataset with more elaborate ground-truth iconographic captions.

%
%
%

\bibliographystyle{splncs04}
\bibliography{references}

\begin{thebibliography}{10}
\providecommand{\url}[1]{\texttt{#1}}
\providecommand{\urlprefix}{URL }
\providecommand{\doi}[1]{https://doi.org/#1}

\bibitem{baraldi2018aligning}
Baraldi, L., Cornia, M., Grana, C., Cucchiara, R.: Aligning text and document
  illustrations: towards visually explainable digital humanities. In: 2018 24th
  International Conference on Pattern Recognition (ICPR). pp. 1097--1102. IEEE
  (2018)

\bibitem{bongini2020visual}
Bongini, P., Becattini, F., Bagdanov, A.D., Del~Bimbo, A.: Visual question
  answering for cultural heritage. arXiv preprint arXiv:2003.09853  (2020)

\bibitem{castellano2020towards}
Castellano, G., Vessio, G.: Towards a tool for visual link retrieval and
  knowledge discovery in painting datasets. In: Italian Research Conference on
  Digital Libraries. pp. 105--110. Springer (2020)

\bibitem{cetinic2018fine}
Cetinic, E., Lipic, T., Grgic, S.: Fine-tuning convolutional neural networks
  for fine art classification. Expert Systems with Applications  \textbf{114},
  107--118 (2018)

\bibitem{cetinic2019deep}
Cetinic, E., Lipic, T., Grgic, S.: A deep learning perspective on beauty,
  sentiment, and remembrance of art. IEEE Access  \textbf{7},  73694--73710
  (2019)

\bibitem{cetinic2020learning}
Cetinic, E., Lipic, T., Grgic, S.: Learning the principles of art history with
  convolutional neural networks. Pattern Recognition Letters  \textbf{129},
  56--62 (2020)

\bibitem{chen2019uniter}
Chen, Y.C., Li, L., Yu, L., Kholy, A.E., Ahmed, F., Gan, Z., Cheng, Y., Liu,
  J.: Uniter: Learning universal image-text representations. arXiv preprint
  arXiv:1909.11740  (2019)

\bibitem{cornia2020explaining}
Cornia, M., Stefanini, M., Baraldi, L., Corsini, M., Cucchiara, R.: Explaining
  digital humanities by aligning images and textual descriptions. Pattern
  Recognition Letters  \textbf{129},  166--172 (2020)

\bibitem{couprie1983iconclass}
Couprie, L.D.: Iconclass: an iconographic classification system. Art Libraries
  Journal  \textbf{8}(2),  32--49 (1983)

\bibitem{crowley2014search}
Crowley, E.J., Zisserman, A.: In search of art. In: European Conference on
  Computer Vision. pp. 54--70. Springer (2014)

\bibitem{deng2020exploring}
Deng, Y., Tang, F., Dong, W., Ma, C., Huang, F., Deussen, O., Xu, C.: Exploring
  the representativity of art paintings. IEEE Transactions on Multimedia
  (2020)

\bibitem{denkowski2014meteor}
Denkowski, M., Lavie, A.: Meteor universal: Language specific translation
  evaluation for any target language. In: Proceedings of the ninth workshop on
  statistical machine translation. pp. 376--380 (2014)

\bibitem{devlin2018bert}
Devlin, J., Chang, M.W., Lee, K., Toutanova, K.: Bert: Pre-training of deep
  bidirectional transformers for language understanding. arXiv preprint
  arXiv:1810.04805  (2018)

\bibitem{elgammal2018shape}
Elgammal, A., Liu, B., Kim, D., Elhoseiny, M., Mazzone, M.: The shape of art
  history in the eyes of the machine. In: 32nd AAAI Conference on Artificial
  Intelligence, AAAI 2018. pp. 2183--2191. AAAI press (2018)

\bibitem{garcia2018read}
Garcia, N., Vogiatzis, G.: How to read paintings: semantic art understanding
  with multi-modal retrieval. In: Proceedings of the European Conference on
  Computer Vision (ECCV). pp.~0--0 (2018)

\bibitem{garcia2020dataset}
Garcia, N., Ye, C., Liu, Z., Hu, Q., Otani, M., Chu, C., Nakashima, Y.,
  Mitamura, T.: A dataset and baselines for visual question answering on art.
  arXiv preprint arXiv:2008.12520  (2020)

\bibitem{guptatowards}
Gupta, J., Madhu, P., Kosti, R., Bell, P., Maier, A., Christlein, V.: Towards
  image caption generation for art historical data. AI methods for digital
  heritage, Workshop at KI2020 43rd German Conference on Artificial
  Intelligence  (2020)

\bibitem{hayn2017subjective}
Hayn-Leichsenring, G.U., Lehmann, T., Redies, C.: Subjective ratings of beauty
  and aesthetics: correlations with statistical image properties in western oil
  paintings. i-Perception  \textbf{8}(3),  2041669517715474 (2017)

\bibitem{jenicek2019linking}
Jenicek, T., Chum, O.: Linking art through human poses. In: 2019 International
  Conference on Document Analysis and Recognition (ICDAR). pp. 1338--1345. IEEE
  (2019)

\bibitem{krishna2017visual}
Krishna, R., Zhu, Y., Groth, O., Johnson, J., Hata, K., Kravitz, J., Chen, S.,
  Kalantidis, Y., Li, L.J., Shamma, D.A., et~al.: Visual genome: Connecting
  language and vision using crowdsourced dense image annotations. International
  journal of computer vision  \textbf{123}(1),  32--73 (2017)

\bibitem{lin2004rouge}
Lin, C.Y.: Rouge: A package for automatic evaluation of summaries. In: Text
  summarization branches out. pp. 74--81 (2004)

\bibitem{lin2014microsoft}
Lin, T.Y., Maire, M., Belongie, S., Hays, J., Perona, P., Ramanan, D.,
  Doll{\'a}r, P., Zitnick, C.L.: Microsoft coco: Common objects in context. In:
  European conference on computer vision. pp. 740--755. Springer (2014)

\bibitem{lu2019vilbert}
Lu, J., Batra, D., Parikh, D., Lee, S.: Vilbert: Pretraining task-agnostic
  visiolinguistic representations for vision-and-language tasks. In: Advances
  in Neural Information Processing Systems. pp. 13--23 (2019)

\bibitem{madhu2019recognizing}
Madhu, P., Kosti, R., M{\"u}hrenberg, L., Bell, P., Maier, A., Christlein, V.:
  Recognizing characters in art history using deep learning. In: Proceedings of
  the 1st Workshop on Structuring and Understanding of Multimedia heritAge
  Contents. pp. 15--22 (2019)

\bibitem{panofsky1972studies}
Panofsky, E.: Studies in iconology. humanistic themes in the art of the
  renaissance, new york. New York: Harper and Row  (1972)

\bibitem{papineni2002bleu}
Papineni, K., Roukos, S., Ward, T., Zhu, W.J.: Bleu: a method for automatic
  evaluation of machine translation. In: Proceedings of the 40th annual meeting
  of the Association for Computational Linguistics. pp. 311--318 (2002)

\bibitem{posthumus2020brill}
Posthumus, E.: Brill iconclass ai test set  (2020)

\bibitem{ren2015faster}
Ren, S., He, K., Girshick, R., Sun, J.: Faster r-cnn: Towards real-time object
  detection with region proposal networks. In: Advances in neural information
  processing systems. pp. 91--99 (2015)

\bibitem{sandoval2019two}
Sandoval, C., Pirogova, E., Lech, M.: Two-stage deep learning approach to the
  classification of fine-art paintings. IEEE Access  \textbf{7},  41770--41781
  (2019)

\bibitem{sargentis2020aesthetical}
Sargentis, G., Dimitriadis, P., Koutsoyiannis, D., et~al.: Aesthetical issues
  of leonardo da vinci’s and pablo picasso’s paintings with stochastic
  evaluation. Heritage  \textbf{3}(2),  283--305 (2020)

\bibitem{seguin2016visual}
Seguin, B., Striolo, C., Kaplan, F., et~al.: Visual link retrieval in a
  database of paintings. In: European Conference on Computer Vision. pp.
  753--767. Springer (2016)

\bibitem{sharma2018conceptual}
Sharma, P., Ding, N., Goodman, S., Soricut, R.: Conceptual captions: A cleaned,
  hypernymed, image alt-text dataset for automatic image captioning. In:
  Proceedings of the 56th Annual Meeting of the Association for Computational
  Linguistics (Volume 1: Long Papers). pp. 2556--2565 (2018)

\bibitem{shen2019discovering}
Shen, X., Efros, A.A., Aubry, M.: Discovering visual patterns in art
  collections with spatially-consistent feature learning. In: Proceedings of
  the IEEE Conference on Computer Vision and Pattern Recognition. pp.
  9278--9287 (2019)

\bibitem{sheng2019generating}
Sheng, S., Moens, M.F.: Generating captions for images of ancient artworks. In:
  Proceedings of the 27th ACM International Conference on Multimedia. pp.
  2478--2486 (2019)

\bibitem{stefanini2019artpedia}
Stefanini, M., Cornia, M., Baraldi, L., Corsini, M., Cucchiara, R.: Artpedia: A
  new visual-semantic dataset with visual and contextual sentences in the
  artistic domain. In: International Conference on Image Analysis and
  Processing. pp. 729--740. Springer (2019)

\bibitem{strezoski2018omniart}
Strezoski, G., Worring, M.: Omniart: a large-scale artistic benchmark. ACM
  Transactions on Multimedia Computing, Communications, and Applications (TOMM)
   \textbf{14}(4),  1--21 (2018)

\bibitem{tan2019lxmert}
Tan, H., Bansal, M.: Lxmert: Learning cross-modality encoder representations
  from transformers. arXiv preprint arXiv:1908.07490  (2019)

\bibitem{vedantam2015cider}
Vedantam, R., Lawrence~Zitnick, C., Parikh, D.: Cider: Consensus-based image
  description evaluation. In: Proceedings of the IEEE conference on computer
  vision and pattern recognition. pp. 4566--4575 (2015)

\bibitem{vinyals2015show}
Vinyals, O., Toshev, A., Bengio, S., Erhan, D.: Show and tell: A neural image
  caption generator. In: Proceedings of the IEEE conference on computer vision
  and pattern recognition. pp. 3156--3164 (2015)

\bibitem{xia2020xgpt}
Xia, Q., Huang, H., Duan, N., Zhang, D., Ji, L., Sui, Z., Cui, E., Bharti, T.,
  Zhou, M.: Xgpt: Cross-modal generative pre-training for image captioning.
  arXiv preprint arXiv:2003.01473  (2020)

\bibitem{young2014image}
Young, P., Lai, A., Hodosh, M., Hockenmaier, J.: From image descriptions to
  visual denotations: New similarity metrics for semantic inference over event
  descriptions. Transactions of the Association for Computational Linguistics
  \textbf{2},  67--78 (2014)

\bibitem{zhou2020unified}
Zhou, L., Palangi, H., Zhang, L., Hu, H., Corso, J.J., Gao, J.: Unified
  vision-language pre-training for image captioning and vqa. In: AAAI. pp.
  13041--13049 (2020)

\end{thebibliography}
%
%
%
%
%

\end{document}